\documentclass[11pt]{article}

\usepackage{acl}

\usepackage{times}
\usepackage{latexsym}

\usepackage[T1]{fontenc}

\usepackage[utf8]{inputenc}

\usepackage{microtype}

\usepackage{inconsolata}

\usepackage{graphicx}
\usepackage{amsmath}
\usepackage{amssymb}
\usepackage{booktabs}
\usepackage{multirow}
\usepackage{tabularx}
\usepackage{xcolor}
\usepackage{subcaption}

\usepackage{soul}

\title{No-Worse Context-Aware Decoding: Preventing Neutral Regression in Context-Conditioned Generation}

\author{Yufei Tao \\
  Affiliation / Address line 1 \\
  Affiliation / Address line 2 \\
  Affiliation / Address line 3 \\
  \texttt{email@domain} \\\And
  Ameeta Agrawal \\
  Affiliation / Address line 1 \\
  Affiliation / Address line 2 \\
  Affiliation / Address line 3 \\
  \texttt{email@domain} \\}

 \author{
Yufei Tao \qquad Ameeta Agrawal \\
Department of Computer Science, Portland State University, USA \\
\texttt{\{yutao, ameeta\}@pdx.edu}
}

\begin{document}
\maketitle

\begin{abstract}
Large language models (LLMs) can answer questions and summarize documents when conditioned on external contexts (e.g., retrieved evidence), yet context use remains unreliable: models may overwrite an already-correct output (\emph{neutral regression}) even when the context is non-informative. We formalize neutral regression as a do-no-harm requirement and quantify it by measuring accuracy drops on baseline-correct items under answer-consistent contexts. We propose No-Worse Context-Aware Decoding (NWCAD), a decode-time adapter built on a two-stream setup with a two-stage gate: it backs off to no-context decoding when the context is non-informative, and otherwise uses context-conditioned decoding with a CAD-style fallback under uncertainty. We evaluate NWCAD on benchmarks that separate do-no-harm reliability from \emph{context utilization} (accuracy gains on genuinely helpful contexts). NWCAD prevents neutral regression on baseline-correct items while preserving strong context-driven accuracy on helpful contexts.
\end{abstract}

\section{Introduction}

Large language models (LLMs) can answer questions and generate summaries when given external contexts, but context-conditioned generation is not automatically reliable. In many context-conditioned workflows like retrieval-augmented generation (RAG)~\cite{lewis2020rag,izacard2021fid} and user-provided context, the provided passage is often \emph{partially} relevant: it may mention related entities, type-matched facts, or near-miss numbers, even when it does not actually entail the correct answer. In such situations, conditioning on context can subtly shift the model's next-token distribution and cause an avoidable change in the final output.

We call this failure mode \emph{neutral regression}: the model overwrites an already-correct answer even though the context is effectively non-informative. Neutral regression is easy to miss in standard evaluations under aggregate accuracy, motivating evaluations that separate neutral and helpful-context cases. Symmetrically, we also care about context utilization: when the context is genuinely helpful, a decoder should be able to use it rather than over-trusting the model's parametric answer.

\begin{figure*}
    \centering
    \includegraphics[width=\linewidth]{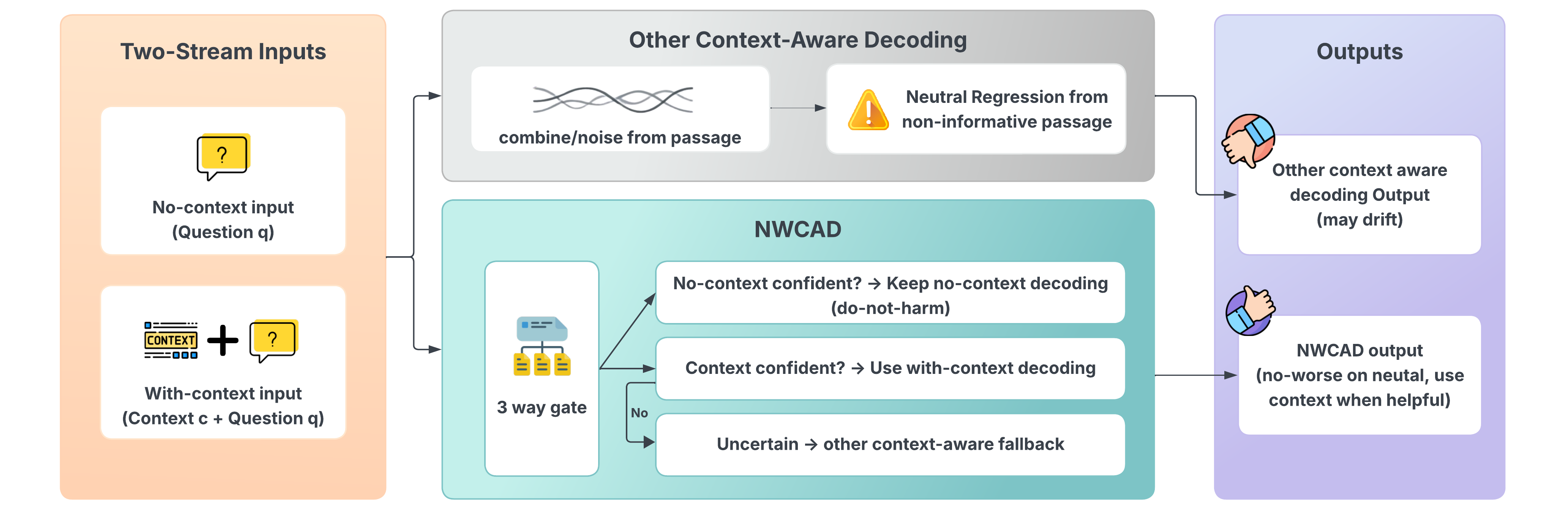}
    \caption{Overview of NWCAD showing two stream inputs when the no-context stream is confident, it keeps the no-context decision (preventing neutral regression); when the with-context stream is confident, it uses with-context decoding; otherwise it invokes a CAD-style fallback decoder.} 
    \label{fig:overview}
\end{figure*}



A common remedy is context-aware decoding: contrast the model's distribution with and without the context, and \emph{tilt} generation toward tokens boosted by context (e.g., CAD and adaptive variants such as AdaCAD and CoCoA~\cite{shi2023trusting,wang2025adacad,khandelwal2025cocoa}).
While effective on average in conflict cases, these decoders lack a do-no-harm guarantee: even when the base model is correct and the context provides weak or noisy evidence, a small distributional shift can change a token choice and cascade into a different (and sometimes wrong) answer. 

Because standard benchmarks mix neutral and conflict contexts, neutral regressions can be obscured by averaged accuracy; we therefore evaluate separately on (i) questions that are already answered correctly without context (do-no-harm) and (ii) questions where the context can correct the answer (\emph{context utilization}). This separation also makes the core trade-off explicit: when the context is non-informative, a decoder should preserve the no-context output; when it is informative, it should shift toward context to correct the answer. A practical decoder should therefore behave like an explicit adapter with an exact no-context backoff, rather than always applying a logit tilt. 
This exact backoff capability is a qualitative difference from continuous logit-tilting decoders: if a method always perturbs logits, it cannot guarantee reproducing the no-context output on neutral inputs.

To address this do-no-harm vs.\ context utilization tension, we propose \textbf{No-Worse Context-Aware Decoding (NWCAD)}, a decode-time adapter built on a two-stream setup with a two-stage gate that yields a three-way routing decision (Figure~\ref{fig:overview} presents an overview): it first decides when to ignore the context and keep the no-context decoding (preventing neutral regression), and when it does use the context, it routes between standard context-conditioned decoding and a contrastive decoder when the context signal is uncertain. Our main contributions are as follows: 
\begin{itemize}
    \item systematically characterizing neutral regression and building controlled evaluations that separate do-no-harm reliability from context utilization; 
    \item proposing NWCAD\footnote{\url{https://github.com/CastGryff/NWCAD}}, a two-stage gate that provably preserves the no-context output whenever it selects the backoff branch (i.e., sets $z'^t=z_0^t$) under greedy decoding; 
    \item evaluating NWCAD across multiple models and a diverse QA benchmark suite to test generalizability; and 
    {\item showing empirically that most context-aware decoding largely reduces to \emph{regime selection} between no-context and with-context decoding, with contrastive mixing rarely needed.}
\end{itemize}

\section{Related Work}

\paragraph{Retrieval-augmented generation.}
Retrieval-augmented generation (RAG) couples context retrieval with conditional generation to improve factuality and access up-to-date information~\cite{lewis2020rag,izacard2021fid}. Recent work highlights that RAG systems can be brittle to distractor evidence and context dilution, motivating context selection and fixed-budget evidence assembly~\cite{li2024knowledgeselection,lahmy2025replacedontexpand,iratni2025dynamiccontext}.

\paragraph{Contextual grounding and verification.}
A broad line of work studies how to make generation faithful to provided evidence, including fine-grained evaluation of factuality/faithfulness and lightweight verification against grounding documents~\cite{maynez2020faithfulness,kryscinski2020evaluating,min-etal-2023-factscore,zhang-etal-2024-fine,tang2024minicheckefficientfactcheckingllms,lost_in_the_later}. Recent RAG-centric frameworks further emphasize context-faithful behavior under real retrieval noise~\cite{nguyen2024sfrragcontextuallyfaithfulllms}.

\paragraph{Selective answering and risk control.}
Selective prediction/abstention trades off risk against coverage by refusing to answer (or filtering generations) when uncertainty is high~\cite{tomani2024uncertaintyabstention,nie2024facttest,wang2025safer}. These methods typically operate at the response level (e.g., abstain or sample-then-filter) and provide complementary mechanisms for controlling errors under uncertainty.

\paragraph{Context-aware decoding.}
A separate line of work modifies decoding directly by contrasting with-context and no-context distributions. {Context-Aware Decoding (CAD)}~\cite{shi2023trusting} biases toward tokens that become more likely when context is present. {Adaptive CAD (AdaCAD)}~\cite{wang2025adacad} varies the tilt strength with the divergence between the two distributions, reducing but not eliminating over-correction. {CoCoA}~\cite{khandelwal2025cocoa} adds confidence signals (e.g., entropy and margin/peakedness) to modulate tilt dynamically. These methods improve average factuality, yet they share two limitations. First, they offer \emph{no formal guarantee of non-regression}: even on neutral inputs where the base model was already correct and the context adds no new information, they can still alter the output. Second, their continuous logit reweighting can still flip token choices in low-conflict settings, even when context provides little new information.

NWCAD is a decode-time adapter for context-conditioned generation that targets neutral regression. Relative to response-level abstention or verifier-based approaches, it requires no additional models and operates directly on the token distribution. Relative to CAD/AdaCAD/CoCoA-style two-stream context-aware decoding, it adds an explicit backoff to no-context decoding on low-divergence steps, yielding a per-step no-neutral-regression under greedy decoding and making the do-no-harm vs.\ context utilization trade-off tunable via simple thresholds.

\section{Methodology}
In this section we formalize our problem setting and introduce {No-Worse Context-Aware Decoding (NWCAD)}. 

\subsection{Setup and Definitions}
We assume a left-to-right autoregressive language model with vocabulary $\mathcal{V}$ and consider generation given a fixed prompt (e.g., a question) and an optional context (e.g., a retrieved passage). At decoding step $t$, both the with-context and no-context streams share the same generated prefix, but differ in whether the external context is included.

Let $z_c^t \in \mathbb{R}^{|\mathcal{V}|}$ be the logits obtained by conditioning on the \emph{contextual input} (context + prompt + prefix), and let $z_0^t$ be the logits obtained by conditioning on the \emph{context-free input} (prompt + prefix only). We denote the corresponding next-token distributions by
\[
p_c^t = \text{softmax}(z_c^t), \quad p_0^t = \text{softmax}(z_0^t).
\]

We use two lightweight signals to decide when to trust each stream: (i) a \emph{context pressure} score based on divergence, and (ii) a simple \emph{confidence} score based on the top-1 margin. 

For divergence, let $D^t$ denote the Jensen-Shannon divergence between the two token distributions at step $t$, $D^t = \text{JS}(p_c^t \,\|\, p_0^t)$. Since computing JS over the full vocabulary is expensive, we approximate it over the union of the top-$K$ tokens from both streams (we use $K{=}50$).  This top-$K$ approximation closely matches full-vocab JS and does not change neutrality/backoff decisions on a representative QA sample (See Appendix~\ref{sec:js_topk_sanity}). We use low divergence as a proxy for ``neutral'' steps (context not materially changing the next-token decision) and high divergence as ``conflict'' steps (context exerting pressure):
\[
\text{Neutral}(t) \;=\; \{\, D^t \leq \tau \,\}.
\]

$D^t$ measures how much the context changes the model's next-token distribution. When $D^t$ is small, the two streams agree; this is consistent with a non-informative context, but it is not sufficient on its own (e.g., both streams may be uncertain yet similar). We therefore combine divergence with a confidence margin before treating a step as safe to back off.

For confidence, we use the top-1 margin of a distribution, i.e., the probability gap between its most likely and second-most likely tokens. We denote this margin by $p_1-p_2$ and use it as a proxy for ``how decisive'' a stream is at a given step. Margins let us distinguish agreement with high confidence (safe backoff) from agreement under uncertainty (route to the context side).

\subsection{From Context-Aware Decoding to No-Worse Decoding}
Context-aware decoding methods such as CAD, AdaCAD, and CoCoA combine a context-free stream and a with-context stream by \emph{tilting} toward tokens that become more likely when context is present. A representative form is
\[
z_{\text{CAD}}^t \;=\; z_c^t \;+\; \alpha \big(z_c^t - z_0^t\big) \;=\; (1+\alpha)z_c^t - \alpha z_0^t,
\]
with a fixed $\alpha$ (CAD) or an adaptive schedule $\alpha^t$ based on divergence or confidence signals (AdaCAD, CoCoA). These methods can improve conflict cases on average, but they introduce a key failure mode: \emph{neutral regression}. Even when the context adds no useful information, the tilt is still applied, so small distributional differences can change a token choice and cascade into a different sequence.

In contrast, NWCAD is a adapter rather than a new tilting rule: it can exactly back off to the no-context stream on neutral steps and otherwise routes to either standard with-context decoding or a CAD-style fallback decoder. NWCAD addresses neutral regression by backing off to no-context decoding when the context appears non-informative, while still allowing context-driven corrections when the context is informative.

The neutrality threshold $\tau$ controls this trade-off: larger $\tau$ makes the decoder more conservative (more backoff to the no-context stream), while smaller $\tau$ increases context utilization.
Unlike continuous tilting (which always perturbs logits), an explicit gate can copy $z_0^t$ exactly; under greedy decoding, this prevents ``small'' context pressure from flipping an argmax token and cascading into a different answer.

\subsection{No-Worse Context-Aware Decoding (NWCAD)}
NWCAD maintains two parallel forward passes (with context vs.\ without) and uses a two-stage gate to choose which logits to decode from at each step.

\paragraph{Stage 1 (BC gate; baseline-correct): exact backoff to the no-context stream.}
When the two distributions agree ($D^t \le \tau$) and the no-context stream is decisive ($p_{0,1}^t-p_{0,2}^t \ge \kappa_{\text{pri}}$), we copy logits from the no-context stream: $z'^t=z_0^t$.
Under greedy decoding, whenever Stage~1 applies and we set $z'^t=z_0^t$, the next token is identical to the no-context stream at step $t$ (immediate since $\arg\max_i z'^t_i=\arg\max_i z_{0,i}^t$). Consequently, if Stage~1 applies at every step, the full decoded sequence matches the no-context output exactly.

\paragraph{Stage 2 (CC gate; context-confident): use context when decisive; fallback otherwise.}
If Stage~1 does not apply, we route to the context side. If the with-context stream is decisive ($p_{c,1}^t-p_{c,2}^t \ge \kappa_{\text{ctx}}$), we decode from $z_c^t$; otherwise we decode from a plug-in CAD-style fallback decoder $z_{\text{fallback}}^t$.

\paragraph{Decision flow.}
At each token, NWCAD either (i) preserves the no-context decision (Stage~1), or (ii) uses the context stream when it is confident, and only falls back to a stronger contrastive decoder on a small set of uncertain steps (Stage~2).

The per-step logic can be summarized as:
\[
z'^t \;=\;
\begin{cases}
z_0^t, & D^t \le \tau \ \land\ (p_{0,1}^t-p_{0,2}^t) \ge \kappa_{\text{pri}} \\
z_c^t, & (p_{c,1}^t-p_{c,2}^t) \ge \kappa_{\text{ctx}} \\
z_{\text{fallback}}^t, & \text{CAD-style fallback} 
\end{cases}
\]

Stage~2 starts only when Stage~1 does not select the no-context stream. At that point, NWCAD either uses standard with-context decoding if the with-context stream is confident, or it uses the CAD-style fallback otherwise. In the ablations below, we therefore distinguish NWCAD$_{\text{BC}}$ (Stage~1 only) which is a no-fallback variant and the full two-stage NWCAD (Stage~1 BC gate + Stage~2 CC gate) with a plug-in fallback decoder; unless stated otherwise, the fallback is CoCoA. 
We default to CoCoA because it reports strong conflict-setting performance in its original study; however, the fallback is plug-in and can be replaced (e.g., CAD/AdaCAD), which we report as ablations. We use {NWCAD}$_{\text{BC}}$ for the Stage~1-only ablation (BC gate only), and {NWCAD}$_X$ when instantiating the fallback decoder as a specific two-stream decoder $X$ (e.g., CAD/AdaCAD/CoCoA).

\begin{table}[!t]
\centering
\small
\begin{tabularx}{\linewidth}{@{}p{0.22\linewidth}X@{}}
\toprule
\textbf{Question} & When were the Articles of Confederation put into effect? \\
\textbf{Gold answer} & March 1, 1781 \\
\midrule
\textbf{Restated context} & The Articles of Confederation officially went into effect on March 1, 1781, marking the first governing framework for the newly independent states.  \\
\textbf{Distractor context} & The Articles of Confederation were officially adopted after several years of debate among the thirteen states. Although they were drafted in 1777, it took until 1780 for all states to ratify the articles and put the new confederation government into operation. \\
\bottomrule
\end{tabularx}
\caption{Example restated and distractor contexts from augmented NQ-open. Distractors add type-matched pressure without entailing the gold answer.} 
\label{tab:nq_examples}
\end{table}


\section{Part I: Controlled Evaluation on Augmented NQ-open}
\label{sec:part1}
Part I uses controlled datasets to diagnose neutral regression and tune a small set of gating thresholds on this diagnostic benchmark. We then freeze these thresholds and evaluate on full-slice QA and beyond-QA tasks in Part II.

\subsection{Augmented NQ-open Benchmark}
\label{sec:datasets}

We start from NQ-open \cite{lee-etal-2019-latent}, an open-domain QA dataset with short answers grounded in Wikipedia. For each question, we attach two types of additional context:
(1) \textbf{restated} contexts, which restate the gold fact and are answer-consistent, and
(2) \textbf{distractor} contexts, which introduce type-matched but incorrect information (e.g., a nearby year), creating plausible but misleading context.
These slices are designed to test whether models avoid regressions when context is not helpful. Table~\ref{tab:nq_examples} provides an example. We also construct a \textbf{helpful} slice by selecting questions that the base model answers incorrectly and adding restated contexts intended to help. This results in 651 restated, 763 distractor, and 637 helpful examples in the full pool.

We automatically verify that restated and helpful contexts contain the gold short answer span (100\% of released examples), and that distractor contexts exclude the gold answer after SQuAD normalization (99.3\%) while still including the intended type-matched distractor (98.5\%). 

Why use controlled subsets? On the full dataset, changes in a model’s answer are hard to interpret: the baseline may already be wrong, or the context may be incorrect or only weakly relevant, so it is unclear whether a change is an improvement or a regression. To avoid this ambiguity, we evaluate on controlled subsets where the desired behavior is clear. For the restated and distractor settings, we use only \emph{baseline-correct} examples, where the no-context baseline matches the gold answer with high confidence (min\_prefix\_margin $\ge 0.8$ over the first 8 tokens). In this case, the correct behavior is to preserve the baseline answer. For the helpful setting, we use examples where the baseline is wrong but standard decoding with context is correct, so improvements directly reflect better use of context. In total, we evaluate on 900 examples, with 300 in each subset.

\subsection{Models and decoding setup}
We evaluate on three open-weight instruction-tuned models: Llama-3.1-8B-Instruct and Llama-3.1-70B-Instruct~\cite{llama3modelcard}, and Ministral-3-8B-Instruct-2512~\cite{mistral3}. We use greedy decoding with max\_new\_tokens=32 for all methods to isolate decode-time effects. For two-stream methods, we run both a with-context and no-context forward pass; NWCAD uses a top-$K$ union approximation ($K\!=\!50$) for JS divergence and uses CoCoA as the fallback decoder in our main results (i.e., NWCAD$_{\text{CoCoA}}$). We tune the thresholds on Llama-3.1-8B controlled slices and then \emph{reuse} the same settings for Ministral-3-8B and Llama-3.1-70B (more details on the tuning experiments are included in the Appendix Table~\ref{tab:nwcad_transfer_summary}.

To test whether neutral regression generalizes beyond open-weight models, we additionally evaluate No-context vs.\ With-context on two API models (\texttt{gpt-5-mini-2025-08-07} and \texttt{gpt-5.2-2025-12-11}) under deterministic decoding; since token-level logits are not exposed, we do not run NWCAD/CAD-style decoders for these API models.

\begin{figure}[!t]
    \centering
    \includegraphics[width=0.99\linewidth]{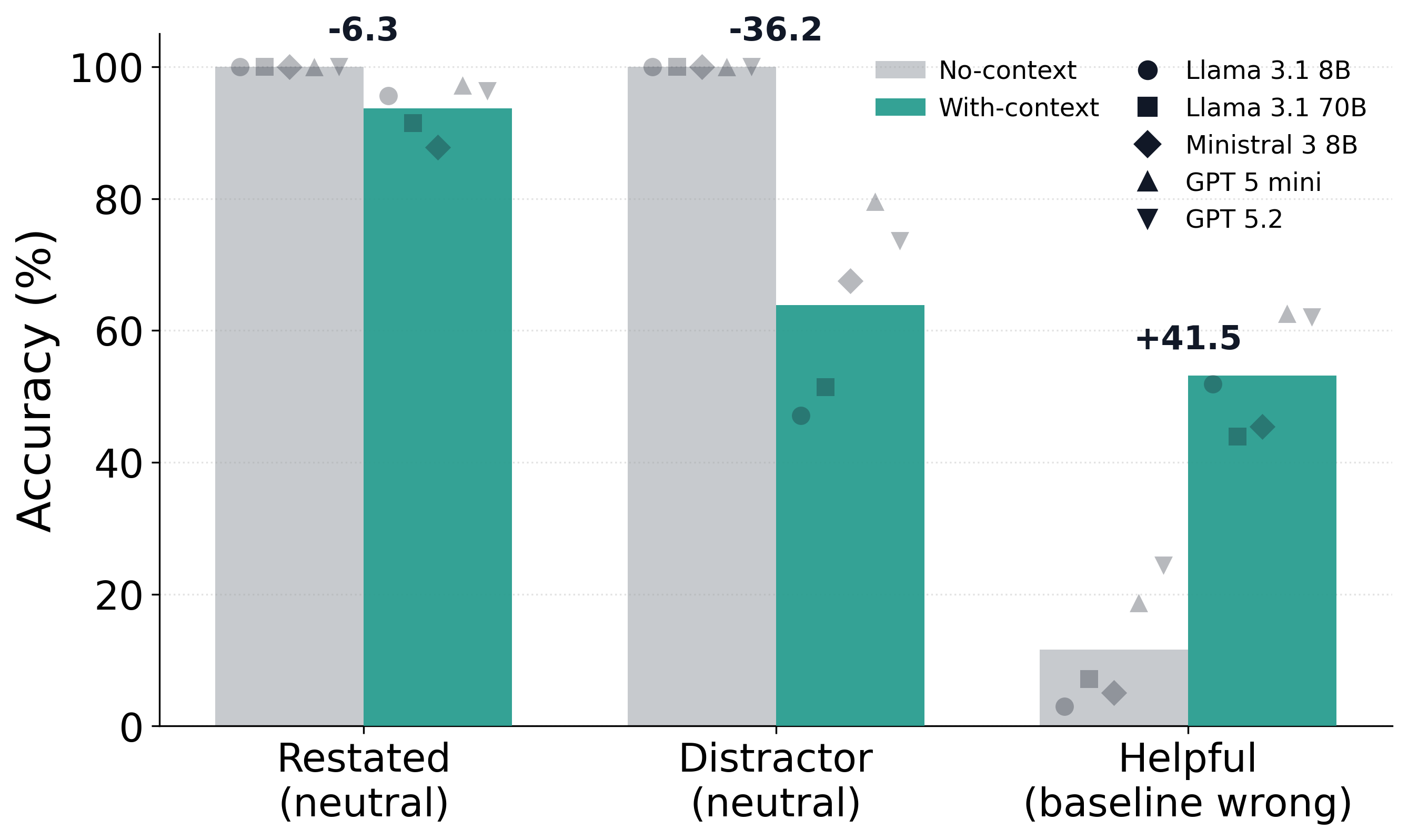}
    \caption{QA accuracy (no-context and with-context) across models and three controlled slices. Averaged across models, adding context decreases accuracy on neutral (Restated/Distractor) but increases accuracy on helpful examples where the baseline is wrong.} 
    \label{fig:neutral_regression_all_models}
\end{figure}

\subsection{Baseline methods}
We compare decoding without context (no-context), standard with-context decoding (with-context), and context-aware decoders (CAD, AdaCAD, CoCoA), alongside NWCAD. 

\subsection{Metrics}
On QA we report exact-match accuracy (SQuAD-normalized), which evaluates short-answer matches after standard normalization (lowercasing and removing punctuation and articles).
We compute EM on a short answer extracted from the model output; Appendix~\ref{app:prompting_setup} details the prompting and extraction protocol.
For controlled-slice summaries, we use a micro-averaged (count-weighted) average across slices.

\begin{figure*}[!t]
    \centering
    \includegraphics[width=0.85\linewidth]{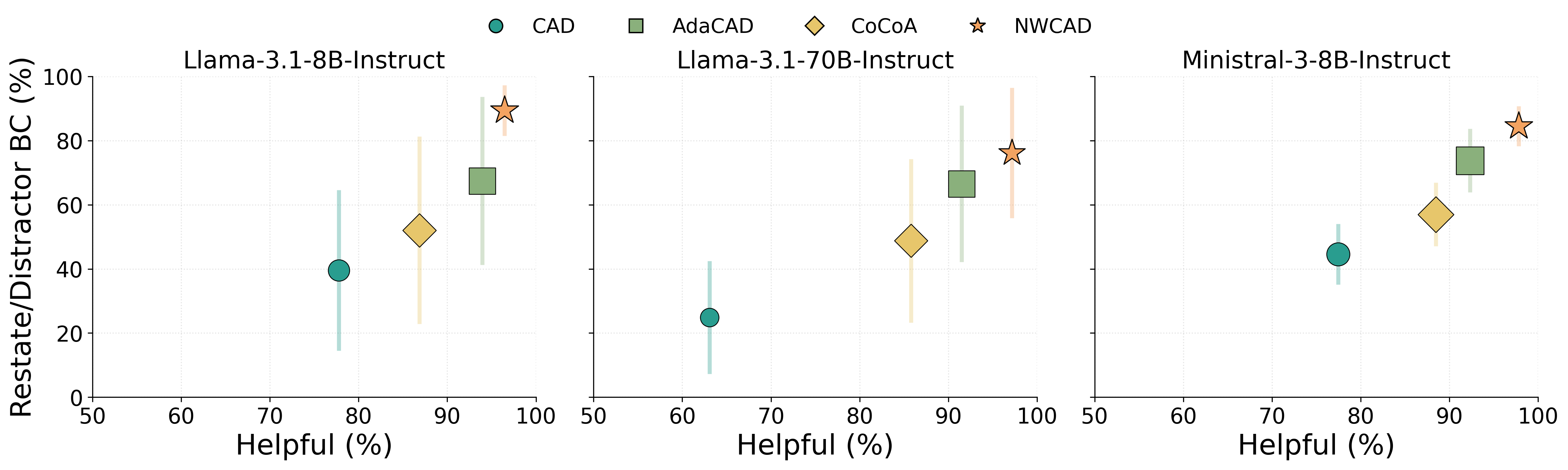}
    \caption{Controlled QA tradeoff between neutral preservation and context utilization ({accuracy}; \%). Each panel plots Helpful accuracy (x-axis) against neutrality on BC subsets (y-axis; marker at mean of Restate/Distractor, vertical bar spans the two). NWCAD moves up and right across all models, improving both neutrality under distractors and helpful gains on helpful contexts.}
    \label{fig:nwr}
\end{figure*}

\begin{table*}[!t]
\centering
\small
\begin{tabularx}{\textwidth}{@{}p{2.1cm}X@{}}
\toprule
\textbf{Neutral (Restated)} &
\textbf{Q:} who played dumbledore after the first one died \newline
\textbf{Context (excerpt):} After the first Dumbledore actor died, Michael Gambon became known for continuing the role in the same magical films. \newline
\textbf{Gold/Base:} Michael Gambon \quad
\textbf{With-context:} Michael Gambon \quad
\textbf{CAD:} Richard Harris \quad
\textbf{AdaCAD:} Richard Harris \quad
\textbf{CoCoA:} Richard Harris \quad
\textbf{NWCAD:} Michael Gambon \\
\midrule
\textbf{Neutral (Distractor)} &
\textbf{Q:} when did the us normalize relations with china \newline
\textbf{Context (excerpt):} Although the process culminated in a major announcement in 1978, it reflected decades of evolving U.S.-China relations. \newline
\textbf{Gold/Base:} January 1, 1979 \quad
\textbf{With-context:} 1978 \quad
\textbf{CAD:} 1978 \quad
\textbf{AdaCAD:} 1978 \quad
\textbf{CoCoA:} 1978 \quad
\textbf{NWCAD:} January 1, 1979 \\
\midrule
\textbf{Helpful-context (NQ-open)} &
\textbf{Q:} who is the voice of the t rex in the good dinosaur \newline
\textbf{Context (excerpt):} Samuel Pack Elliott is an American actor recognized for his deep sonorous voice. In the animated film featuring a T Rex, Sam Elliott's distinctive tone would fit the prehistoric world as naturally as another veteran performer’s voice might in a similar dinosaur adventure. \newline
\textbf{Gold:} Sam Elliott \quad
\textbf{Base (no-context):} Arlen Ness \quad
\textbf{With-context:} Sam Elliott \quad
\textbf{CAD/AdaCAD/CoCoA:} Woody Harrelson \quad
\textbf{NWCAD:} Sam Elliott \\
\bottomrule
\end{tabularx}
\caption{Qualitative examples illustrating neutral regression and context correction. NWCAD preserves baseline-correct answers on neutral contexts and uses context when it is helpful.} 
\label{tab:qual}
\end{table*}

\subsection{Results and Discussion}
We organize results around the do-no-harm vs.\ context utilization tension: preserve baseline-correct answers when the context is non-informative, while leveraging context when it is helpful.

\paragraph{Neutral regression across model families.}
We first validate that neutral regression is not specific to any single model family or to CAD-style contrastive decoding. Figure~\ref{fig:neutral_regression_all_models} compares the no-context baseline against standard with-context decoding on baseline-correct (BC) neutral subsets: since BC filtering ensures the baseline matches the gold short answer, baseline achieves {100\% accuracy} by construction; any drop under with-context decoding reflects an avoidable regression induced by a not-helpful (answer-consistent) context.
Across three open-source models and two API models, with-context decoding exhibits consistent regressions especially under distractor-hard contexts (middle bar), while simultaneously improving accuracy on helpful contexts (rightmost bar). This empirical trade-off motivates NWCAD's decode-time gating.


\paragraph{Separating do-no-harm from context utilization.}
Figure~\ref{fig:nwr} reports {accuracy} on controlled QA slices that isolate the two objectives. Across all evaluated open-weight models, NWCAD achieves the best weighted average by improving both neutral preservation and context utilization. In particular, continuous tilting methods appear to over-react to distractor contexts (large drops on distractor-hard BC), whereas NWCAD's explicit backoff prevents these avoidable regressions while still allowing corrections on the helpful subset.

\begin{figure*}[!t]
    \centering
    \includegraphics[width=1\linewidth]{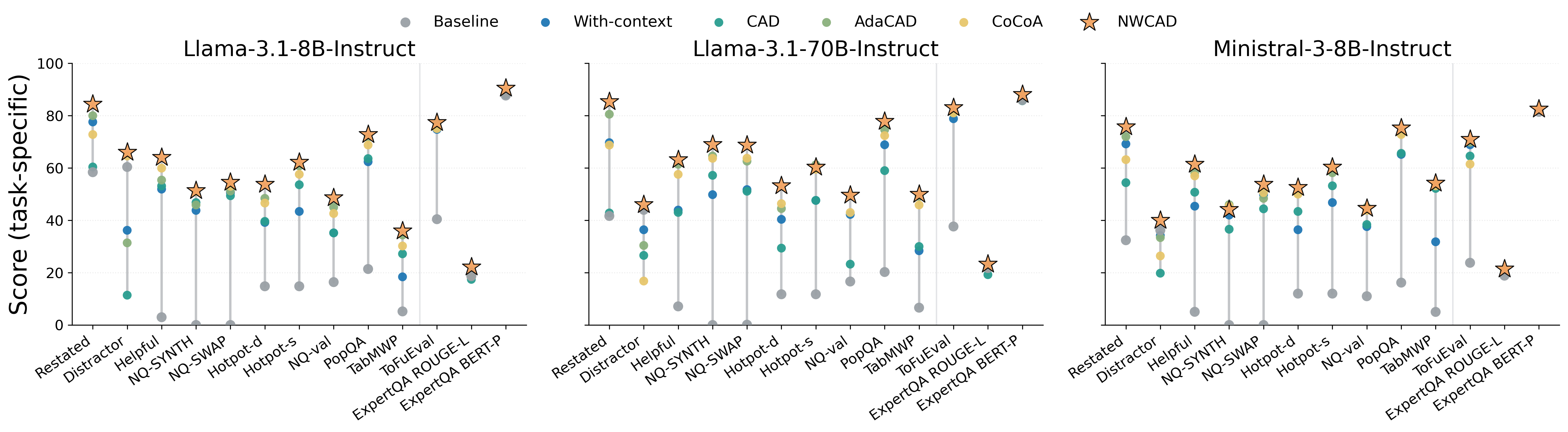}
    \caption{Full-slice results across QA and beyond-QA benchmarks (higher is better). Restate/Distractor/Helpful through TabMWP report EM; ToFuEval uses AlignScore; ExpertQA reports ROUGE-L and BERTScore-P. {NQ-SYNTH/NQ-SWAP are context-defined, so near-zero Baseline accuracy is expected}. For each task, the vertical bar shows the Baseline$\rightarrow$NWCAD score lift, with other methods plotted for reference.}
    \label{fig:qa_general}
\end{figure*}

\subsection{Qualitative Analysis}
Table~\ref{tab:qual} illustrates some representative cases from our augmented NQ-open benchmark. The first two examples show \emph{neutral regression}: a context contains type-matched but non-decisive information (e.g., a related year), and CAD-style tilting drifts away from a correct base answer. The final example shows the \emph{helpful-context} setting, where the context is genuinely informative and NWCAD allows a correction.

\section{Part II: Full-slice Evaluation}
\label{sec:part2}
Part II evaluates on full-slice benchmarks that mix neutral/conflict and retrieval noise. We use the same models, decoding setup, and metrics as in Part~I unless otherwise noted.

\subsection{Benchmarks}
\paragraph{Full-slice QA suite.}
After diagnosing and tuning on controlled slices, we evaluate on 12 full slice QA benchmarks that mix neutral and conflict contexts. We report exact match QA accuracy on NQ SYNTH and NQ SWAP with synthetic or swapped contexts \cite{wang2025adacad,khandelwal2025cocoa}, HotpotQA distractor and HotpotQA support which are multi hop QA settings with distractor versus supporting retrieval contexts \cite{yang-etal-2018-hotpotqa}, NQ val short which is the original NQ open validation short answer set, PopQA which is an open domain QA benchmark \cite{mallen-etal-2023-trust}, and TabMWP which evaluates table as context math word problems \cite{lu2023tabmwp}. For efficiency, we use 1,000 example subsets for NQ SWAP and HotpotQA. Following the paired no context and with context prompt design used by prior context-aware decoding work such as CAD AdaCAD and CoCoA, we evaluate all open weight decoders on identical prompt pairs for each dataset example and vary only the decode time rule.

\paragraph{Beyond QA.} 
To test whether NWCAD generalizes beyond short answer QA, we additionally evaluate on two more datasets: ToFuEval (topic-focused dialogue summarization)~\cite{tang-etal-2024-tofueval} and ExpertQA (expert-curated long-form answers)~\cite{malaviya-etal-2024-expertqa}, reporting AlignScore on ToFuEval and ROUGE-L/BERTScore-P on ExpertQA.

\subsection{Results and Discussion}
{Figure~\ref{fig:qa_general} summarizes the main results 
(Appendix~\ref{app:full_results} reports the full per-dataset numbers)}. We observe that NWCAD improves robustness on the augmented slices while remaining competitive (and often improving) on the diverse QA benchmarks. {The largest improvements occur in distractor heavy settings such as Distractor hard and HotpotQA distractor, consistent with the neutrality gate preventing type matched but non decisive context from overriding correct answers. Gains are not limited to conservative backoff: NWCAD also improves Helpful and context defined tasks (NQ SYNTH and NQ SWAP), indicating the adapter still exploits context when the context stream is confident.} 
{Because exact match can be overly strict for short free-form answers, we also evaluate the QA outputs from all three open-weight backbones with GPT-4o-mini as an LLM judge (Appendix~\ref{app:llm_judge}). In the updated semantic-evaluation table, the same overall conclusion still holds: NWCAD is the strongest overall method across the three backbones, with especially clear gains on Restate-hard and Distractor-hard. This indicates that the improvements are not simply an artifact of strict EM scoring.}


{Across all three models, NWCAD also achieves the best ToFuEval AlignScore and improves ExpertQA ROUGE-L/BERTScore-P over both With-context and contrastive decoding baselines, suggesting the adapter generalizes beyond short-answer QA.} 

\section{Ablations}
We report additional analyses on component ablations (NWCAD$_{\text{BC}}$ vs. full NWCAD), NWCAD as an adapter over CAD/AdaCAD/CoCoA, routing frequency (how often the fallback decoder is used), and latency relative to the corresponding base decoder.

\begin{figure}[!t]
    \centering
    \includegraphics[width=.9\linewidth]{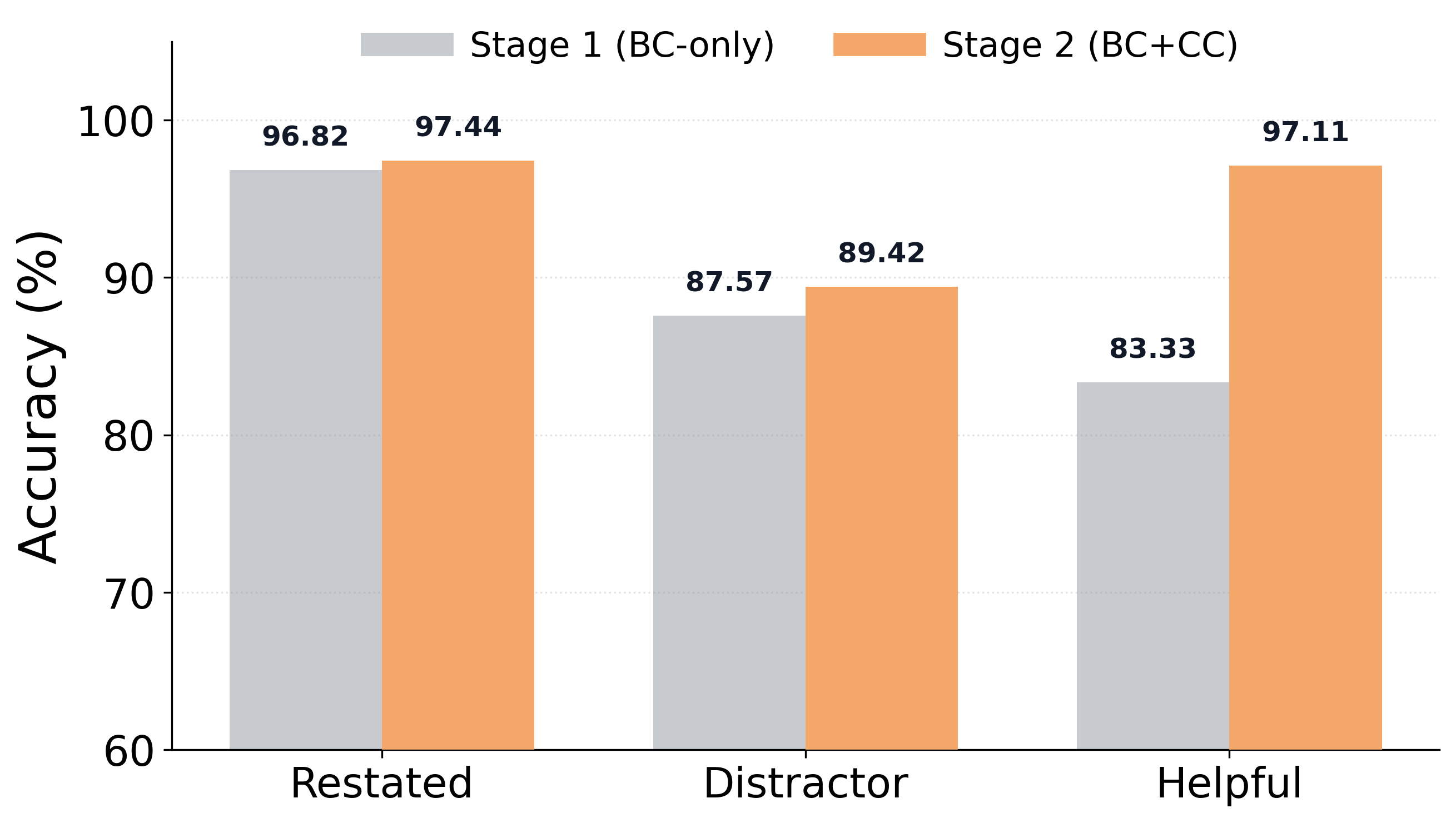}
    \caption{NWCAD$_{\text{BC}}$ vs.\ NWCAD.}
    \label{fig:component_ablation_improvement}
\end{figure}

\begin{figure}[!t]
    \centering
    \includegraphics[width=1\linewidth]{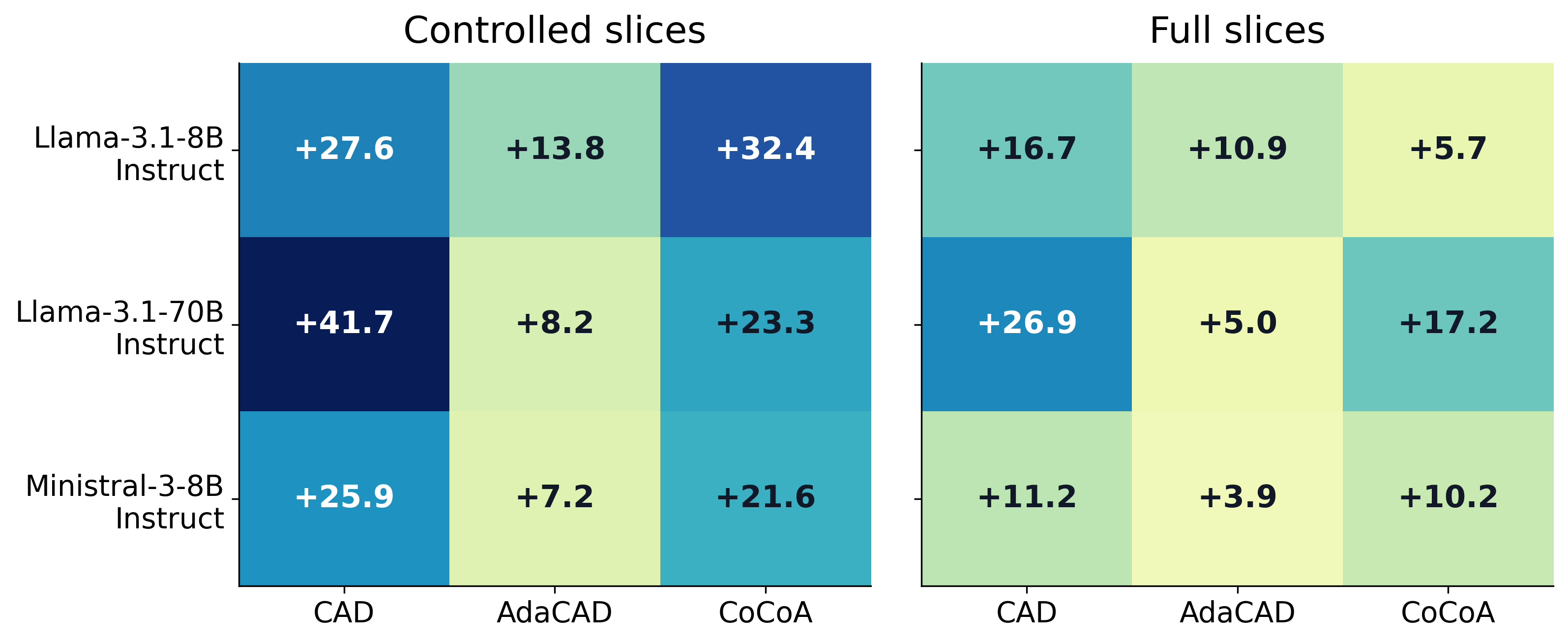}
    \caption{NWCAD as an adapter over existing decoders. Each cell shows the gain from wrapping a base decoder $X$ with NWCAD$_X$ across three backbones, on controlled-slice weighted average  and full-slice average.}
    
    \label{fig:nwcad_adapter}
\end{figure}

\paragraph{NWCAD$_{\text{BC}}$ vs.\ NWCAD.}
{NWCAD was developed sequentially: NWCAD$_{\text{BC}}$ uses Stage~1 (BC gate) to protect baseline-correct neutral cases (tuning $\tau$ to control conservatism), and the full method adds Stage~2 (CC gate + CAD-style fallback) to better utilize clearly helpful contexts. Figure~\ref{fig:component_ablation_improvement} shows that Stage~2 improves performance on all testing cases on average by 5.2\% (detailed results in Appendix~\ref{app:component_analysis}). 
{This shows that Stage~2 mainly helps when the model should use the context; the no-fallback ablation below further suggests that most of the gain comes from choosing between no-context and standard with-context decoding, while the CAD-style fallback adds a smaller extra benefit.}}

\paragraph{NWCAD as an adapter over CAD/AdaCAD/CoCoA.}

NWCAD is a adapter that can be layered on top of different two stream decoders by swapping the fallback implementation. For a base decoder $X$, we instantiate {NWCAD}$_X$ by using $X$ as the fallback decoder in Stage~2. Figure~\ref{fig:nwcad_adapter} show that NWCAD consistently improves over each corresponding base decoder across models (Appendix~\ref{app:adapter_tables} reports full results). On controlled slices, gains range from about 7 to over 40 accuracy points, with the largest improvements observed when wrapping CAD and CoCoA, which are more susceptible to neutral regression under distractor contexts. Improvements are driven primarily by large boosts on Restated and Distractor subsets, while Helpful accuracy is preserved or modestly improved, indicating that NWCAD’s explicit backoff and routing primarily correct overreaction to weak or noisy context rather than suppressing genuine context utilization.

\paragraph{No-fallback ablation.}
{Table~\ref{tab:nwcad_nofallback} reports a no-fallback ablation on a 100-example-per-slice subset. To isolate the effect of the routing decision itself, we evaluate a simplified \emph{No-fallback} variant: if Stage~1 does not back off to the no-context stream, decoding routes directly to standard with-context logits, without using the CAD-style fallback. Here, NWCAD$_{\text{BC}}$ denotes the Stage~1-only variant, \emph{No-fallback} denotes the no-fallback variant, and NWCAD denotes the full two-stage method. The result stays very close to full NWCAD, which suggests that most of the gain comes from choosing the right branch, while the CAD-style fallback adds a smaller extra benefit.}

\begin{table}[!t]
\centering
\small
\setlength{\tabcolsep}{2pt}
\resizebox{\linewidth}{!}{%
\begin{tabular}{p{1.4cm}ccccc}
\toprule
Slice & No-ctx & With-ctx & NWCAD$_{\text{BC}}$ & No-fallback & NWCAD \\
\midrule
Restate-hard & 48 & 83 & 80 & \textbf{85} & \textbf{85} \\
Distractor-hard & \textbf{50} & 29 & 31 & 31 & 31 \\
Helpful & 8 & \textbf{64} & 52 & 62 & 62 \\
NQ-SWAP & 0 & \textbf{52} & \textbf{52} & \textbf{52} & 51 \\
\bottomrule
\end{tabular}
}
\caption{{No-fallback ablation (in \%) on a 100-example-per-slice subset (Llama-3.1-8B; accuracy; \%). Here, NWCAD$_{\text{BC}}$ denotes the Stage~1-only variant, while \emph{No-fallback} routes directly to standard with-context decoding whenever Stage~1 does not back off, without using the CAD-style fallback.}}
\label{tab:nwcad_nofallback}
\end{table}

\paragraph{Routing statistics (how often the fallback is used).}
The fallback decoder is intended as a safety net for a small set of uncertain steps, not the default path. As shown in Table~\ref{tab:NWCAD_routing_stats}, it is used for only about 1-2\% of generated tokens across tasks, which means NWCAD’s improvements do not come from frequent fallback use. Instead, NWCAD mainly switches between the no-context and with-context streams on a per-token basis, using the fallback only occasionally. This suggests that the key to effective context-aware decoding is deciding {\em when} to trust the model’s prior knowledge versus the provided context, rather than constantly blending the two at every step. On context-driven benchmarks (e.g., NQ-SWAP), routing toward the context stream dominates, while fallback use remains minimal.

\begin{table}[!t]
\setlength{\tabcolsep}{2pt}
\centering
\small
\begin{tabular}{p{1.8cm}cccc}
\toprule
Slice & No-Context & Context & Fallback & Any-fallback \\
\midrule
Restate-hard & 75.14 & 23.65 & 1.21 & 5.68 \\
Distractor-hard & 73.70 & 24.35 & 1.95 & 8.65 \\
Helpful & 63.06 & 35.06 & 1.88 & 10.68 \\
\midrule
NQ-SYNTH & 49.57 & 48.66 & 1.78 & 6.70 \\
NQ-SWAP & 37.88 & 60.26 & 1.86 & 7.10 \\
HotpotQA-distractor & 60.53 & 38.39 & 1.08 & 4.70 \\
\bottomrule
\end{tabular}
\caption{Routing frequencies for NWCAD's three-way decision (\%). (\emph{No-Context} = BC/no-context, \emph{Context} = CC/with-context, \emph{Fallback} = CAD-style fallback, \emph{Any-fallback} is the fraction of examples that use the fallback at least once). The CAD-style fallback is rarely invoked. } 
\label{tab:NWCAD_routing_stats}
\end{table}

\paragraph{Efficiency and Latency.}
{Because NWCAD introduces additional control logic and, in principle, extra computation, it is important to verify that the reliability gains do not come at a inference cost.}
We report Llama-3.1-8B-Instruct relative decoding latency (sec/token) of NWCAD$_X$ versus its base decoder $X$ on a single RTX 5090 GPU (FP16; microbatch=1; 30 examples per workload) From the results presented in Table~\ref{tab:latency}, we observe that across ToFuEval and ExpertQA, NWCAD stays within about 2\% of $X$; on short-answer QA it can be faster because the adapter often routes to the prior/context streams and invokes contrastive mixing only rarely.

\begin{table}[!t]
\centering
\small
\setlength{\tabcolsep}{2pt}
\begin{tabular}{lccc}
\toprule
Pair (NWCAD$_X$/X) & QA mean & ToFuEval  & ExpertQA  \\
\midrule
NWCAD$_{\text{CAD}}$/CAD & 0.88 & 1.01 & 0.99 \\
NWCAD$_{\text{AdaCAD}}$/AdaCAD & 0.99 & 1.01 & 0.98 \\
NWCAD$_{\text{CoCoA}}$/CoCoA & 0.90 & 0.98 & 1.01 \\
\bottomrule
\end{tabular}
\caption{Relative decoding latency (sec/token; ratio) of NWCAD$_X$ to its base decoder $X$. Values below 1 mean NWCAD is faster than the base model. NWCAD is either similar or slightly faster.} 
\label{tab:latency}
\end{table}




\section{Conclusion}
We introduced No-Worse Context-Aware Decoding as a decode-time mechanism for context-conditioned generation. NWCAD addresses {neutral regression} by preserving the no-context decision while still shifting toward context under conflict. 
Across all tests, NWCAD consistently improves over CAD, AdaCAD and CoCoA style decoders by reducing regressions under distractor contexts while preserving gains on genuinely helpful contexts. More broadly, our results point to \emph{regime selection} as a useful abstraction for future context-aware decoding: for most steps, generation is best handled by either the no-context stream or the context stream, with contrastive mixing needed only rarely. This perspective motivates conditional-compute adapters that dynamically decide when and how to rely on contextual evidence, rather than continuously mixing streams at every step. Looking forward, key directions include extending the neutral regression prevention beyond greedy decoding and improve reliability for long form outputs. 

\section*{Limitations}
{Our work is not without limitations. First, our default thresholds are tuned on one model (Llama-3.1-8B) and transferred to other models and benchmarks; while this works well in our experiments, these values may not be optimal for new models, domains, or prompting setups.} Second, NWCAD requires token-level access to both no-context and with-context logits, so it does not directly apply to black-box API models that do not expose logits.

\section*{Acknowledgments}
We sincerely thank the anonymous reviewers and the members of the PortNLP group for their valuable feedback and helpful suggestions.


\bibliography{latex/custom}

\appendix

\numberwithin{table}{section}
\numberwithin{figure}{section}

\setcounter{section}{0}
\setcounter{table}{0}
\setcounter{figure}{0}



\section{Top-$K$ JS Approximation Validation}
\label{sec:js_topk_sanity}
We approximate JS divergence over the union of the top-$K$ tokens from both streams ($K{=}50$ in the main paper). To assess whether this approximation affects routing decisions, we compare it against full-vocab JS on Llama-3.1-8B-Instruct using 50 examples per controlled slice (150 total; 784 generated tokens; greedy decoding with max\_new\_tokens=32, $\tau{=}0.3$, $\kappa_{\text{pri}}{=}0.30$). We also checked $K{=}200$ and observed identical $\text{Neutral}(t)$ and Stage~1 (BC gate) backoff decisions with even smaller JS error (Table~\ref{tab:js_topk_sanity}).

\begin{table*}[!t]
\centering
\small
\begin{tabular}{lccccc}
\toprule
Slice & Tokens & Neutral flip (\%) & Stage~1 flip (\%) & JS MAE & JS corr \\
\midrule
Restate-hard & 219 & 0.00 & 0.00 & 0.0003 & 0.9999 \\
Distractor-hard & 242 & 0.00 & 0.00 & 0.0012 & 0.9999 \\
Helpful & 323 & 0.00 & 0.00 & 0.0012 & 0.9999 \\
\bottomrule
\end{tabular}
\caption{Top-$K$ union JS ($K{=}50$) vs.\ full-vocab JS on controlled-slice decoding steps. The approximation yields zero token-step disagreements for neutrality/backoff decisions and closely matches full-vocab JS (MAE $\le$ 0.0012).}
\label{tab:js_topk_sanity}
\end{table*}

\section{NWCAD Parameter Tuning}
\label{app:nwcad_tuning}
\paragraph{Tuning objective and trade-offs.}
NWCAD introduces a small number of interpretable scalars that expose a do-no-harm vs.\ context utilization trade-off at decode time. Increasing the neutrality threshold $\tau$ expands the region where we defer to the no-context stream (improving no-worse behavior on neutral inputs) but can over-gate context utilization on helpful-context inputs. Similarly, higher confidence thresholds $\kappa_{\text{pri}}$ and $\kappa_{\text{ctx}}$ make the two-stage gate more conservative in when it trusts a stream.

\paragraph{Parameters.}
We tune three scalars: the neutrality threshold $\tau$, and the top-1 margin thresholds $\kappa_{\text{pri}}$ and $\kappa_{\text{ctx}}$ used by the BC and CC gates, respectively. Unless otherwise noted, we use the same decoding setup as the main QA experiments (greedy, max\_new\_tokens=32). We select these default thresholds on the Llama-3.1-8B controlled slices (no separate held-out dev split) and then reuse them for all other models and full-slice benchmarks.
{Table~\ref{tab:nwcad_transfer_summary} summarizes this directly. We do \emph{not} tune thresholds separately for each model; instead, we tune them on Llama-3.1-8B and reuse the same settings on Llama-3.1-70B and Ministral-3-8B.}

\begin{table}[t]
\centering
\scriptsize
\setlength{\tabcolsep}{3pt}
\begin{tabular}{lccccc}
\toprule
Model & Source & $\tau$ & $\kappa_{\text{pri}}$ & $\kappa_{\text{ctx}}$ & QA macro \\
\midrule
Llama-3.1-8B & tuned on 8B & 0.30 & 0.30 & 0.05 & 59.39 \\
Llama-3.1-70B & transferred from 8B & 0.30 & 0.30 & 0.05 & 62.35 \\
Ministral-3-8B & transferred from 8B & 0.30 & 0.30 & 0.05 & 56.23 \\
\bottomrule
\end{tabular}
\caption{{Reuse of the same NWCAD thresholds across model families. The defaults are tuned once on Llama-3.1-8B and then reused unchanged on Llama-3.1-70B and Ministral-3-8B. QA macro reports the mean across the ten full-slice QA benchmarks in Table~\ref{tab:qa_general}.}}
\label{tab:nwcad_transfer_summary}
\end{table}

\paragraph{$\tau$ sweep (\do-no-harm vs.\ context utilization frontier).}
To make the do-no-harm vs.\ context utilization trade-off explicit, we sweep $\tau$ on the same model and decoding setup (Table~\ref{tab:NWCAD_tau_sweep}). Restate/Distractor report accuracy on baseline-correct neutral subsets; since the baseline is 100\% by construction on BC, we also report the corresponding accuracy drop ($100-\text{accuracy}$) in parentheses. Helpful reports accuracy on the helpful-context subset; and Overall is the micro-averaged score across these subsets.

\noindent We use {$\tau{=}0.3$} in the main results, since it achieves the best overall accuracy on this combined benchmark while maintaining strong neutral preservation; larger $\tau$ over-gates helpful contexts.

\paragraph{Sensitivity of $\kappa_{\text{pri}}$ (BC gate margin threshold).}
$\kappa_{\text{pri}}$ controls how confident the no-context stream must be for the BC gate to apply on low-divergence steps. With $\tau{=}0.3$ and $\kappa_{\text{ctx}}=0.05$ fixed, we sweep $\kappa_{\text{pri}} \in \{0.00,0.05,0.10,0.15,0.20,0.30\}$ and observe smooth improvements with diminishing returns as $\kappa_{\text{pri}}$ increases (Table~\ref{tab:NWCAD_kappa_pri_sweep}). We use {$\kappa_{\text{pri}}{=}0.30$} in all experiments.

\paragraph{Tuning $\kappa_{\text{ctx}}$ (CC gate).}
With $\tau{=}0.3$ and $\kappa_{\text{pri}}{=}0.30$ fixed, we sweep the context-stream margin threshold $\kappa_{\text{ctx}}$ and report controlled-slice {accuracy} on the baseline-correct (BC) subsets (Table~\ref{tab:triage_ctx_sweep_bc}). We choose $\kappa_{\text{ctx}}{=}0.05$ among the tested values:

\begin{table}[t]
\centering
\small
\begin{tabular}{lcccc}
\toprule
$\tau$ & Restate$\uparrow$ & Distractor$\uparrow$ & Helpful$\uparrow$ & Overall$\uparrow$ \\
\midrule
0.1 & 94.59 & 78.57 & 90.15 & 86.53 \\
0.2 & 95.86 & 83.60  & 87.12 & 88.83 \\
{0.3} & {96.82}& {87.57}  & {83.33} & {90.41} \\
0.5 & 99.04  & 94.44  & 45.45 & 88.35 \\
\bottomrule
\end{tabular}
\caption{Sensitivity of the neutrality threshold $\tau$ on neutral preservation and helpful utilization. $\tau$ exposes a do-no-harm vs.\ context utilization trade-off.}
\label{tab:NWCAD_tau_sweep}
\end{table}

\begin{table}[!t]
\centering
\small
\begin{tabular}{lccc}
\toprule
$\kappa_{\text{pri}}$ & Helpful$\uparrow$ & Restate-hard$\uparrow$ & Distractor-hard$\uparrow$ \\
\midrule
0.00 & 61.42 & 77.11 & 29.36 \\
0.05 & 63.20 & 78.03 & 29.36 \\
0.10 & 63.50 & 78.34 & 29.36 \\
0.15 & 64.39 & 78.80 & 29.62 \\
0.20 & 65.58 & 79.42 & 29.75 \\
{0.30} & {65.58} & {79.57} & {30.01} \\
\bottomrule
\end{tabular}
\caption{Sensitivity sweep for the BC-gate margin threshold $\kappa_{\text{pri}}$ (accuracy; \%). Performance improves smoothly and largely converges by $\kappa_{\text{pri}}{\approx}0.2$-0.3.}
\label{tab:NWCAD_kappa_pri_sweep}
\end{table}

\begin{table}[!t]
\centering
\small
\begin{tabular}{lcccc}
\toprule
$\kappa_{\text{ctx}}$ & Restate$\uparrow$ & Distractor$\uparrow$ & Helpful$\uparrow$ & Combined$\uparrow$ \\
\midrule
0 & 96 & 59 & 71 & 74.67 \\
{0.05} & {97.44} & {79.42} & {97.11} & {89.47} \\
0.10 & 97.44 & 79.42 & 96.53 & 89.35 \\
0.20 & 97.44 & 79.42 & 94.80 & 89.00 \\
0.30 & 97.44 & 79.16 & 94.22 & 88.77 \\
\bottomrule
\end{tabular}
\caption{Sensitivity sweep for the CC-gate margin threshold $\kappa_{\text{ctx}}$ ({accuracy}; \%). We select $\kappa_{\text{ctx}}{=}0.05$.}
\label{tab:triage_ctx_sweep_bc}
\end{table}

\section{Exact Prompting and Decoding Setup}
\label{app:prompting_setup}
\paragraph{QA prompting.}
For instruction-tuned models, we use a chat-style prompt with a \emph{system prompt} and a \emph{user prompt}. The system prompt is:
\texttt{Answer the question concisely with just the fact or name. Do not add explanations or extra sentences.}
The user prompt ends with \texttt{Answer:} to encourage short-answer formatting.

\paragraph{No-context vs.\ with-context.}
Each QA example uses the same question under two conditions: a question-only (no-context) user prompt and a with-context user prompt that prepends the retrieved/synthetic context followed by an explicit marker (\texttt{Using only the references listed above, answer the following question:}). This paired-prompt setup is shared across CAD/AdaCAD/CoCoA/NWCAD: Baseline and With-context correspond to using the no-context or with-context prompt, respectively, while two-stream methods compute both distributions under the same shared generated prefix.

\paragraph{Decoding parameters.}
Unless otherwise noted, we use greedy decoding with \texttt{max\_new\_tokens=32}, \texttt{max\_context\_length=4064}, and a fixed seed (\texttt{seed=42}). Two-stream methods compute both the with-context and no-context distributions at each step under the shared generated prefix.
NWCAD uses a top-$K$ union approximation for JS divergence ($K{=}50$). We use $\tau{=}0.3$, $\kappa_{\text{pri}}{=}0.30$, and $\kappa_{\text{ctx}}{=}0.05$; on the context side, if neither stream is confident, we invoke a CAD-style fallback decoder (CoCoA in our main experiments; the fallback can be swapped, e.g., AdaCAD, as an ablation).
For Llama-3.1-70B, we additionally use quantized loading (int8) and microbatching for memory.

\paragraph{Answer extraction and scoring.}
In most cases the generations are already short answers due to the answer-only prompting. When needed (primarily for CoCoA-style decoding under EOS-suppression variants that can produce long continuations), we apply a lightweight extraction that takes the text following the final \texttt{Answer:} marker (if present) or the first non-empty line. We report {exact-match accuracy (SQuAD-normalized)}, which lowercases and strips punctuation, articles, and extra whitespace before matching.

\section{Full-slice QA and Beyond-QA: Full Tables}
\label{app:full_results}
{Tables~\ref{tab:qa_general} and~\ref{tab:beyond_qa} provide the numeric results underlying Figure~\ref{fig:qa_general}.}

\begin{table*}[t]
\centering
\scriptsize
{\setlength{\tabcolsep}{2.3pt}
\begin{tabular}{llcccccccccc}
\toprule
{Model} & Method & {Restate-hard$\uparrow$} & {Distractor-hard$\uparrow$} & {Helpful$\uparrow$} & {NQ-SYNTH$\uparrow$} & {NQ-SWAP$\uparrow$} & {Hotpot$_d$$\uparrow$} & {Hotpot$_s$$\uparrow$} & {NQ-val$\uparrow$} & {PopQA$\uparrow$} & {TabMWP$\uparrow$} \\
\midrule
\multirow{6}{*}{{Llama-3.1-8B}} & Baseline & 58.40 & 60.40 & 2.97 & 0.00 & 0.00 & 14.80 & 14.80 & 16.40 & 21.40 & 5.20 \\
 & With-context & 77.60 & 36.20 & 51.93 & 43.80 & 50.60 & 39.20 & 43.40 & 35.20 & 62.40 & 18.40 \\
 & CAD & 60.40 & 11.40 & 53.12 & 46.80 & 49.40 & 39.60 & 53.60 & 35.20 & 63.60 & 27.20 \\
 & AdaCAD & 80.00 & 31.40 & 55.39 & 45.87 & 51.20 & 48.40 & 60.80 & 45.00 & 71.80 & 34.40 \\
 & CoCoA & 72.80 & 64.60 & 59.94 & 50.98 & 53.00 & 46.60 & 57.60 & 42.60 & 68.80 & 30.20 \\
 & NWCAD & \textbf{84.40} & \textbf{66.00} & \textbf{64.09} & \textbf{51.40} & \textbf{54.60} & \textbf{53.80} & \textbf{62.20} & \textbf{48.60} & \textbf{72.80} & \textbf{36.00} \\
\midrule
\multirow{6}{*}{{Llama-3.1-70B}} & Baseline & 41.67 & 44.00 & 7.12 & 0.00 & 0.12 & 11.75 & 11.75 & 16.62 & 20.25 & 6.62 \\
 & With-context & 69.68 & 36.40 & 43.92 & 49.81 & 51.75 & 40.38 & 47.62 & 42.25 & 68.88 & 28.38 \\
 & CAD & 42.82 & 26.60 & 43.03 & 57.20 & 51.12 & 29.38 & 47.62 & 23.25 & 59.00 & 30.00 \\
 & AdaCAD & 80.56 & 30.40 & 61.42 & 64.59 & 62.62 & 44.50 & \textbf{61.25} & 49.50 & 74.62 & 49.12 \\
 & CoCoA & 68.75 & 16.80 & 57.57 & 63.70 & 63.75 & 46.38 & 59.50 & 43.00 & 72.38 & 45.88 \\
 & NWCAD & \textbf{85.42} & \textbf{46.00} & \textbf{63.20} & \textbf{68.97} & \textbf{68.75} & \textbf{53.28} & 60.38 & \textbf{49.62} & \textbf{77.88} & \textbf{50.00} \\
\midrule
\multirow{6}{*}{{Ministral-3-8B}} & Baseline & 32.40 & 36.20 & 5.04 & 0.00 & 0.00 & 12.00 & 12.00 & 11.00 & 16.20 & 5.00 \\
 & With-context & 69.20 & 34.20 & 45.40 & 42.00 & 49.80 & 36.40 & 46.80 & 37.60 & 65.20 & 31.80 \\
 & CAD & 54.40 & 19.80 & 50.74 & 36.60 & 44.40 & 43.40 & 53.20 & 38.40 & 65.60 & 52.20 \\
 & AdaCAD & 72.00 & 33.40 & 58.16 & \textbf{46.00} & 48.40 & 50.40 & 58.20 & \textbf{45.00} & 73.40 & 53.80 \\
 & CoCoA & 63.20 & 26.40 & 56.97 & 45.40 & 50.60 & 50.00 & 60.20 & 43.80 & 72.80 & 53.20 \\
 & NWCAD & \textbf{75.80} & \textbf{40.00} & \textbf{61.42} & 44.20 & \textbf{53.80} & \textbf{52.60} & \textbf{60.40} & 44.60 & \textbf{75.29} & \textbf{54.20} \\
\bottomrule
\end{tabular}
}
\caption{Full-slice QA results (\%) on augmented NQ-open and a diverse QA suite. NWCAD improves robustness on the augmented slices while remaining competitive on general QA tasks; NQ-SYNTH/NQ-SWAP are context-defined, so near-zero Baseline (no-context) accuracy is expected (With-context sanity check); CoCoA uses EOS-allowed decoding (Appendix~\ref{app:eos_ablation}).}
\label{tab:qa_general}
\end{table*}

\begin{table*}[t]
\centering
\small
{\setlength{\tabcolsep}{3.2pt}
\begin{tabular}{llccc}
\toprule
{Model} & Method & {ToFuEval$\uparrow$} & {ExpertQA (ROUGE-L)$\uparrow$} & {ExpertQA (BERTScore-P)$\uparrow$} \\
\midrule
\multirow{6}{*}{{Llama-3.1-8B}} & Baseline & 40.44 & 18.65 & 87.70 \\
 & With-context & 74.75 & 20.26 & 88.07 \\
 & CAD & 77.40 & 17.44 & 88.43 \\
 & AdaCAD & 75.75 & 19.01 & 88.33 \\
 & CoCoA & 75.09 & 18.67 & 89.02 \\
 & NWCAD & \textbf{77.44} & \textbf{22.23} & \textbf{90.49} \\
\midrule
\multirow{6}{*}{{Llama-3.1-70B}} & Baseline & 37.65 & 21.76 & 85.98 \\
 & With-context & 78.84 & 22.78 & 86.19 \\
 & CAD & 81.04 & 19.23 & 85.69 \\
 & AdaCAD & 81.40 & 22.57 & 87.69 \\
 & CoCoA & 81.04 & 21.63 & 87.00 \\
 & NWCAD & \textbf{83.12} & \textbf{23.34} & \textbf{88.11} \\
\midrule
\multirow{6}{*}{{Ministral-3}} & Baseline & 23.78 & 18.86 & 81.45 \\
 & With-context & 68.88 & 21.30 & 82.46 \\
 & CAD & 64.57 & 20.55 & 82.43 \\
 & AdaCAD & 69.86 & 21.03 & 82.28 \\
 & CoCoA & 61.49 & 20.83 & 82.37 \\
 & NWCAD & \textbf{70.99} & \textbf{21.47} & \textbf{82.54} \\
\bottomrule
\end{tabular}
}
\caption{Beyond-QA results on ToFuEval and ExpertQA. NWCAD remains competitive and improves the overall score across models.}
\label{tab:beyond_qa}
\end{table*}

\section{LLM-as-a-Judge Evaluation}
\label{app:llm_judge}
{To check whether strict exact-match undercounts semantically correct short answers, we evaluate the QA outputs from all three open-weight backbones with GPT-4o-mini as an LLM judge. Table~\ref{tab:qa_llm_judge} mirrors Table~\ref{tab:qa_general}: semantic evaluation raises absolute scores, but the same overall conclusion remains, with NWCAD strongest overall and especially strong on Restate-hard and Distractor-hard.}
\begin{table*}[t]
\centering
\scriptsize
{\setlength{\tabcolsep}{2.3pt}
\begin{tabular}{llcccccccccc}
\toprule
{Model} & Method & {Restate-hard$\uparrow$} & {Distractor-hard$\uparrow$} & {Helpful$\uparrow$} & {NQ-SYNTH$\uparrow$} & {NQ-SWAP$\uparrow$} & {Hotpot$_d$$\uparrow$} & {Hotpot$_s$$\uparrow$} & {NQ-val$\uparrow$} & {PopQA$\uparrow$} & {TabMWP$\uparrow$} \\
\midrule
\multirow{6}{*}{{Llama-3.1-8B}} & Baseline  & 57.76 & 40.81 & 24.33 & 25.50 & 4.50 & 29.20 & 29.10 & 38.01 & 27.25 & 11.90 \\
 & With-context  & 89.55 & 51.25 & 85.46 & 69.60 & 62.20 & 72.10 & 86.40 & 79.23 & 83.81 & \textbf{54.40} \\
 & CAD  & 75.58 & 32.50 & 78.04 & 64.30 & 64.00 & 69.10 & 80.00 & 65.68 & 78.44 & 49.10 \\
 & AdaCAD  & 89.55 & 48.49 & 86.05 & 69.80 & 66.10 & 76.70 & 85.70 & 76.60 & 82.69 & \textbf{54.40} \\
 & CoCoA  & 85.10 & 39.71 & 83.09 & 64.40 & 66.40 & 73.40 & 83.60 & 73.82 & 80.94 & 50.60 \\
 & NWCAD  & \textbf{90.02} & \textbf{52.16} & \textbf{86.16} & \textbf{70.00} & \textbf{67.00} & \textbf{75.80} & \textbf{86.70} & \textbf{79.82} & \textbf{84.12} & \textbf{54.40} \\
\midrule
\multirow{6}{*}{{Llama-3.1-70B}} & Baseline  & 50.00 & 58.40 & 27.00 & 19.25 & 5.88 & 26.62 & 26.75 & 34.62 & 23.75 & 12.62 \\
 & With-context  & 92.80 & 57.80 & 82.80 & 69.50 & 64.75 & 72.75 & \textbf{87.38} & 77.25 & 82.38 & 60.00 \\
 & CAD  & 67.60 & 29.80 & 76.56 & 66.62 & 71.00 & 51.75 & 74.50 & 52.25 & 72.50 & 49.88 \\
 & AdaCAD  & 92.20 & 54.80 & 89.91 & 76.38 & \textbf{73.00} & 73.50 & 86.88 & 76.88 & 85.62 & 60.12 \\
 & CoCoA  & 89.60 & 52.00 & 87.54 & 71.62 & 70.88 & 69.50 & 84.75 & 69.00 & 84.12 & 57.62 \\
 & NWCAD  & \textbf{94.40} & \textbf{62.20} & \textbf{90.21} & \textbf{79.00} & \textbf{73.00} & \textbf{74.38} & 86.25 & \textbf{78.88} & \textbf{87.12} & \textbf{61.00} \\
\midrule
\multirow{6}{*}{{Ministral-3-8B}} & Baseline  & 39.48 & 41.42 & 26.11 & 20.50 & 6.20 & 26.50 & 26.70 & 29.34 & 18.62 & 13.60 \\
 & With-context  & 87.86 & 52.82 & 84.27 & 70.70 & 62.00 & \textbf{79.90} & 87.60 & 79.29 & 85.12 & 63.40 \\
 & CAD  & 78.96 & 41.02 & 80.42 & 64.80 & 63.90 & 76.00 & 85.10 & 71.28 & 80.31 & 62.70 \\
 & AdaCAD  & 86.94 & 53.21 & 83.68 & \textbf{72.80} & 62.10 & 78.90 & 87.80 & 79.17 & 84.94 & \textbf{63.70} \\
 & CoCoA  & 74.19 & 40.50 & 72.70 & 58.90 & 54.60 & 70.40 & 74.50 & 63.60 & 74.00 & 53.40 \\
 & NWCAD  & \textbf{88.33} & \textbf{55.83} & \textbf{85.16} & 71.80 & \textbf{66.90} & \textbf{79.90} & \textbf{88.20} & \textbf{79.95} & \textbf{85.69} & 63.60 \\
\bottomrule
\end{tabular}
}
\caption{{GPT-4o-mini semantic judging on the QA outputs across all three open-weight backbones (accuracy; \%), in the same layout as Table~\ref{tab:qa_general}. For Llama-3.1-70B we evaluate the same subset sizes used in its corresponding QA table (500/800 examples per slice); Llama-3.1-8B and Ministral-3-8B use the full available QA slices. Relative to exact match, semantic evaluation increases absolute scores while keeping the same overall picture, with NWCAD strongest overall and especially strong on distractor-heavy / no-harm-oriented slices.}}
\label{tab:qa_llm_judge}
\end{table*}

\section{NWCAD$_{\text{BC}}$ vs.\ NWCAD.}
\label{app:component_analysis}
Table~\ref{tab:NWCAD_component_ablation} 
{reports the controlled-slice comparison. Together with the no-fallback ablation below, it suggests that Stage~2 mainly helps when the model should use the context, while most of the gain comes from selecting the right branch.}
\begin{table*}[t]
\centering
\small
\begin{tabular}{p{0.45\linewidth}cccc}
\toprule
Setting & Restates BC$\uparrow$ & Distractor BC$\uparrow$ & Helpful$\uparrow$ & Combined$\uparrow$ \\
\midrule
{NWCAD$_{\text{BC}}$}  & 96.82 & 87.57 & 83.33 & 90.41 \\
{NWCAD} & \textbf{97.44} & \textbf{89.42} & \textbf{97.11} & \textbf{94.47} \\
\bottomrule
\end{tabular}
\caption{{Component ablation comparing NWCAD$_{\text{BC}}$ (Stage~1 only) vs.\ full two-stage NWCAD on controlled slices (accuracy; \%).} Stage~2 improves helpful accuracy but can trade off distractor-hard BC accuracy.}
\label{tab:NWCAD_component_ablation}
\end{table*}

{Table~\ref{tab:NWCAD_component_ablation_full} reports the corresponding full-slice QA ablation on Llama-3.1-8B using the same setup as the main QA results. As in the controlled-slice comparison above, NWCAD$_{\text{BC}}$ already captures most of the gain, while the full two-stage method adds smaller additional improvements on top.}
\begin{table*}[t]
\centering
\small
\begin{tabular}{lcccc}
\toprule
Benchmark & Baseline & With-context & NWCAD$_{\text{BC}}$ & NWCAD \\
\midrule
Restated-hard & 58.40 & 77.60 & 82.10 & \textbf{84.40} \\
Distractor-hard & 60.40 & 36.20 & 64.80 & \textbf{66.00} \\
Helpful & 2.97 & 51.93 & 61.00 & \textbf{64.09} \\
NQ-SYNTH & 0.00 & 43.80 & 51.10 & \textbf{51.40} \\
NQ-SWAP & 0.00 & 50.60 & 53.10 & \textbf{54.60} \\
HotpotQA-distractor & 14.80 & 39.20 & 52.00 & \textbf{53.80} \\
HotpotQA-support & 14.80 & 43.40 & 62.00 & \textbf{62.20} \\
NQ-val-short & 16.40 & 35.20 & 46.00 & \textbf{48.60} \\
PopQA & 21.40 & 62.40 & 71.60 & \textbf{72.80} \\
TabMWP & 5.20 & 18.40 & 35.00 & \textbf{36.00} \\
\bottomrule
\end{tabular}
\caption{{Full-slice QA ablation comparing NWCAD$_{\text{BC}}$ (Stage~1 only) vs.\ full NWCAD on Llama-3.1-8B (accuracy; \%). These values match the rebuttal table.}}
\label{tab:NWCAD_component_ablation_full}
\end{table*}

\section{NWCAD as an Adapter: Full Results}
\label{app:adapter_tables}
{Tables~\ref{tab:nwcad_adapter} and~\ref{tab:nwcad_adapter_full} provide the numeric results underlying Figure~\ref{fig:nwcad_adapter}.}

\begin{table*}[t]
\centering
\small
\begin{tabular}{llccc}
\toprule
Model & Base decoder $X$ & Base & {NWCAD$_X$} & $\Delta$ \\
\midrule
\multirow{3}{*}{{Llama-3.1-8B}} & CAD & 46.61 & 74.18 & +27.57 \\
 & AdaCAD & 71.73 & 85.51 & +13.78 \\
 & CoCoA & 58.06 & 90.50 & +32.44 \\
\midrule
\multirow{3}{*}{{Llama-3.1-70B}} & CAD & 34.19 & 75.92 & +41.73 \\
 & AdaCAD & 71.59 & 79.78 & +8.19 \\
 & CoCoA & 57.14 & 80.42 & +23.28 \\
\midrule
\multirow{3}{*}{{Ministral-3}} & CAD & 55.23 & 81.10 & +25.87 \\
 & AdaCAD & 79.63 & 86.79 & +7.16 \\
 & CoCoA & 67.16 & 88.71 & +21.55 \\
\bottomrule
\end{tabular}
\caption{{Controlled-slice weighted-average accuracy (\%) for each base decoder $X$ and its NWCAD$_X$ wrapper (fallback decoder = $X$).} {NWCAD improves over each corresponding base decoder across evaluated models.}}
\label{tab:nwcad_adapter}
\end{table*}

\begin{table*}[t]
\centering
\small
\begin{tabular}{llccc}
\toprule
Model & Base decoder & {$\Delta$ Restate-hard} & {$\Delta$ Distractor-hard} & {$\Delta$ Helpful} \\
\midrule
\multirow{3}{*}{{Llama-3.1-8B}} & CAD & +20.00 & +26.60 & +3.56 \\
 & AdaCAD & +1.00 & +30.80 & +0.99 \\
 & CoCoA & +11.60 & +1.40 & +4.15 \\
\midrule
\multirow{3}{*}{{Llama-3.1-70B}} & CAD & +34.96 & +31.00 & +14.83 \\
 & AdaCAD & +4.39 & +9.80 & +0.89 \\
 & CoCoA & +16.67 & +29.20 & +5.63 \\
\midrule
\multirow{3}{*}{{Ministral-3}} & CAD & +15.80 & +16.00 & +1.78 \\
 & AdaCAD & +3.80 & +5.00 & +2.97 \\
 & CoCoA & +12.60 & +13.60 & +4.45 \\
\bottomrule
\end{tabular}
\caption{Full-slice gains (\%) from wrapping each base decoder $X$ with NWCAD$_X$, computed on the unfiltered augmented NQ-open slices (Restate-hard, Distractor-hard, Helpful). Gains persist on full-slice mixtures.}
\label{tab:nwcad_adapter_full}
\end{table*}

\section{CoCoA EOS Suppression Ablation}
\label{app:eos_ablation}
The official CoCoA release suppresses the EOS token by setting \texttt{--dsab-min-len=5} (released default), which can force long continuations up to \texttt{max\_new\_tokens} and interact with answer extraction in short-answer QA. For short-answer QA, we set \texttt{--dsab-min-len=0} to allow EOS stopping. Table~\ref{tab:cocoa_eos_ablation} reports the resulting performance differences. We report CoCoA with EOS allowed in the main paper and include the EOS-suppressed setting here for reference.

\begin{table*}[t]
\centering
\small
\begin{tabular}{llccc}
\toprule
{Model} & Method & {Restate-hard$\uparrow$} & {Distractor-hard$\uparrow$} & {Helpful$\uparrow$} \\
\midrule
\multirow{4}{*}{{Llama-3.1-8B}} & CoCoA (EOS suppressed) & 72.80 & 17.60 & 59.94 \\
 & CoCoA (EOS allowed) & 72.80 & 64.60 & 59.94 \\
 & NWCAD$_{\text{CoCoA}}$ (EOS suppressed) & 84.40 & 66.00 & 64.09 \\
 & NWCAD$_{\text{CoCoA}}$ (EOS allowed) & 84.40 & 66.00 & 64.09 \\
\midrule
\multirow{4}{*}{{Ministral-3}} & CoCoA (EOS suppressed) & 22.20 & 10.60 & 22.85 \\
 & CoCoA (EOS allowed) & 63.20 & 26.40 & 56.97 \\
 & NWCAD$_{\text{CoCoA}}$ (EOS suppressed) & 75.80 & 40.00 & 61.13 \\
 & NWCAD$_{\text{CoCoA}}$ (EOS allowed) & 75.80 & 40.00 & 61.42 \\
\midrule
\multirow{4}{*}{{Llama-3.1-70B}} & CoCoA (EOS suppressed) & 77.31 & 16.80 & 57.57 \\
 & CoCoA (EOS allowed) & 68.75 & 30.00 & 57.57 \\
 & NWCAD$_{\text{CoCoA}}$ (EOS suppressed) & 85.42 & 46.20 & 63.20 \\
 & NWCAD$_{\text{CoCoA}}$ (EOS allowed) & 85.42 & 46.00 & 63.20 \\
\bottomrule
\end{tabular}
\caption{Effect of CoCoA EOS suppression on the three augmented NQ-open slices (\%). We report CoCoA with EOS allowed in the main paper; the EOS-suppressed setting can be sensitive to model formatting and short-answer extraction.}
\label{tab:cocoa_eos_ablation}
\end{table*}

\section{Full-slice Sensitivity of $\kappa_{\text{ctx}}$}
\label{app:kctx_full_sensitivity}
$\kappa_{\text{ctx}}$ controls Stage~2 routing on the context side: when the context stream is sufficiently confident (top-1 margin $p_1{-}p_2 \ge \kappa_{\text{ctx}}$), NWCAD uses vanilla context logits; otherwise it falls back to a contrastive decoder (CoCoA in our main setup). We tune $\kappa_{\text{ctx}}$ on the controlled baseline-correct (BC) subsets (Table~\ref{tab:triage_ctx_sweep_bc}) to maximize combined BC accuracy. Table~\ref{tab:triage_ctx_sweep_full} repeats the same sweep on full-slice mixtures; results are stable across the tested range, indicating the choice of $\kappa_{\text{ctx}}{=}0.05$ is not brittle.

\begin{table*}[!t]
\centering
\small
{\begin{tabular}{lcccccc}
\toprule
$\kappa_{\text{ctx}}$ & Restate-hard & Distractor-hard & Helpful & NQ-SYNTH & NQ-SWAP & HotpotQA \\
\midrule
0.05 & 75.27 & 46.92 & 57.86 & 58.90 & 53.20 & 54.70 \\
0.10 & 75.12 & 46.53 & 57.57 & 58.70 & 53.30 & 54.30 \\
0.20 & 74.65 & 46.40 & 56.68 & 59.10 & 54.00 & 53.50 \\
0.30 & 74.35 & 46.13 & 56.38 & 58.30 & 53.50 & 53.50 \\
\bottomrule
\end{tabular}}
\caption{Sensitivity of $\kappa_{\text{ctx}}$ on full-slice QA and general datasets ({accuracy}; \%).}
\label{tab:triage_ctx_sweep_full}
\end{table*}

\end{document}